\documentclass[conference]{IEEEtran}
\usepackage{times}

\usepackage[numbers]{natbib}
\usepackage{multicol}
\usepackage[bookmarks=true,hidelinks]{hyperref}

\usepackage{amsmath,amssymb}
\usepackage{graphicx}
\usepackage{booktabs}
\usepackage{array}
\usepackage{xcolor}

\newif\ifanonymousmode
\anonymousmodefalse

\newcommand{\realauthorname}{Jiajie Zhang, Yankai Xiang, Changhao Chen}

\ifanonymousmode

  \newcommand{\paperauthorblock}{\anonauthorname}
  
  \newcommand{\paperpdfauthor}{Anonymous}
  \newcommand{\paperpdfsubject}{Preprint}
\else

  \providecommand{\IEEEoverridecommandlockouts}{}
  \IEEEoverridecommandlockouts

  \newcommand{\paperauthorblock}{%
    Jiajie Zhang$^{1,\dagger}$, Yankai Xiang$^{2,\dagger}$, and Changhao Chen$^{1,*}$%
    \thanks{\raggedright $^{1}$Jiajie Zhang and Changhao Chen are with the Hong Kong
      University of Science and Technology (Guangzhou), Guangzhou, China
      {\tt\small jerryzhangamen7@gmail.com},
      {\tt\small changhaochen@hkust-gz.edu.cn}}%
    \thanks{\raggedright $^{2}$Yankai Xiang is with ShanghaiTech University, Shanghai,
      China {\tt\small xiangyk2025@shanghaitech.edu.cn}}%
    \thanks{$^{\dagger}$These authors contributed equally to this work.}%
    \thanks{$^{*}$Corresponding author.}%
  }
  
  \newcommand{\paperpdfauthor}{\realauthorname}
  \newcommand{\paperpdfsubject}{RSS 2027 Stage 1 draft, author-identifying version}

\fi

\hypersetup{pdfauthor={\paperpdfauthor}, pdftitle={VLAs Execute, Code Recovers: 3D Skill Envelopes Bridge the Compositional Gap in Agentic Manipulation}, pdfsubject={\paperpdfsubject},
  pdfkeywords={manipulation; skill composition; vision-language-action models; agents}}

\newcommand{\NStates}{640}
\newcommand{\NPairs}{64}
\newcommand{\DView}{$+$1.7}
\newcommand{\DWorld}{$+$10.0}
\newcommand{\SDirect}{14.2}
\newcommand{\SHome}{82.8}
\newcommand{\SWorld}{92.8}
\newcommand{\CHome}{17.8}
\newcommand{\CView}{17.0}
\newcommand{\CWorld}{2.0}
\newcommand{\ExDirect}{0/10}
\newcommand{\ExHome}{0/10}
\newcommand{\ExView}{0/10}
\newcommand{\ExWorld}{10/10}
\newcommand{\ExHomeContact}{10 of 10}
\newcommand{\ExViewContact}{10 of 10}
\newcommand{\NBroken}{five}
\newcommand{\NSpared}{three}
\newcommand{\ExBrokenMax}{4/10}
\newcommand{\ExSparedMin}{9/10}
\newcommand{\SAtomic}{98.0}
\newcommand{\DGapHome}{15.2}
\newcommand{\AContact}{0.946}
\newcommand{\ALopo}{0.843}
\newcommand{\ALopoLookup}{0.695}
\newcommand{\DLopoLookupCI}{[$+$0.074, $+$0.215]}
\newcommand{\ALopoNoGeom}{0.499}
\newcommand{\SJRepair}{92.3}
\newcommand{\SJPosthoc}{91.4}
\newcommand{\SJWorld}{92.8}
\newcommand{\DJRepairHomeCI}{[$+$8.3, $+$10.6]}
\newcommand{\NMid}{90}
\newcommand{\SMidHome}{38.9}
\newcommand{\SMidWorld}{93.3}
\newcommand{\SMidRepair}{95.6}
\newcommand{\SMidPosthoc}{84.4}
\newcommand{\NTop}{50}
\newcommand{\STopHome}{42.0}
\newcommand{\STopWorld}{92.0}
\newcommand{\STopRepair}{94.0}
\newcommand{\STopPosthoc}{80.0}
\newcommand{\NNew}{140}
\newcommand{\InvTotalFrac}{1/280}
\newcommand{\PctInv}{0.4}
\newcommand{\PctPrim}{98.6}
\newcommand{\NEnvMin}{37}
\newcommand{\NEnvMax}{89}
\newcommand{\ADepth}{0.944}
\newcommand{\DepthAgree}{140/140}
\newcommand{\AConfLopo}{0.864}
\newcommand{\AConfLopoLookup}{0.605}
\newcommand{\AConfSingle}{0.971}
\newcommand{\AConfSingleLookup}{0.790}
\newcommand{\AConfCross}{0.912}
\newcommand{\AConfCrossLookup}{0.681}
\newcommand{\NChainDrawer}{128}
\newcommand{\SChainBOne}{29.7}
\newcommand{\SChainBTwo}{93.8}
\newcommand{\ChainTwoBOne}{0/32}
\newcommand{\ChainTwoBTwo}{28/32}
\newcommand{\NRQFiveCommon}{429}
\newcommand{\ARQFiveOurs}{0.902}
\newcommand{\ARQFiveBestLang}{0.793}
\newcommand{\DRQFiveBest}{$+$0.110}
\newcommand{\DRQFiveBestCI}{[$+$0.054, $+$0.162]}
\newcommand{\ALangQwenMaxOne}{0.768}
\newcommand{\ALangGlmOne}{0.715}
\newcommand{\ALangGlmTwo}{0.701}
\newcommand{\ALangQwenLocOne}{0.703}
\newcommand{\ALangQwenLocTwo}{0.724}
\newcommand{\AContactEnvEval}{0.961}
\newcommand{\AContactQwenMax}{0.866}
\newcommand{\AStratOurs}{0.834}
\newcommand{\AStratQwenMax}{0.666}
\newcommand{\AStratGlmOne}{0.500}
\newcommand{\SGlmBFour}{91.4}
\newcommand{\NPooledLLM}{256}
\newcommand{\SPooledBThree}{84.0}
\newcommand{\SPooledBFour}{91.4}
\newcommand{\PPooledMcn}{0.0054}
\newcommand{\NQwenMatched}{105}
\newcommand{\SRgbdBFour}{93.8}
\newcommand{\RgbdAgreeSeams}{656/656}
\newcommand{\RgbdMsLow}{341}
\newcommand{\RgbdMsHigh}{368}
\newcommand{\RgbdSkipPct}{87.8}
\newcommand{\NRgbdMatched}{88}
\newcommand{\XSpeedSym}{2.34}
\newcommand{\XSpeedGated}{5.60}
\newcommand{\NBossUnits}{35}
\newcommand{\SBossBThree}{25.7}
\newcommand{\SBossBFour}{60.0}
\newcommand{\DBossDiff}{$+$34.3}
\newcommand{\PBossMcn}{0.0075}
\newcommand{\RepBossCut}{92.2}
\newcommand{\BossMicroBThree}{0/5}
\newcommand{\BossMicroBFour}{3/5}
\newcommand{\NFreeTable}{71}
\newcommand{\NFreeFlag}{20}
\newcommand{\NFreeWine}{23}
\newcommand{\SFreeFlagHome}{45.0}
\newcommand{\SFreeFlagSham}{40.0}
\newcommand{\SFreeFlagOurs}{80.0}
\newcommand{\SFreeFlagWorld}{90.0}
\newcommand{\DFreeFlagCI}{[$+$10.0, $+$60.0]}
\newcommand{\SFreeTabHome}{80.3}
\newcommand{\SFreeTabOurs}{94.4}
\newcommand{\PFreeTabMcn}{0.0063}
\newcommand{\FreeWinePlanPct}{43.5}

\newcommand{\pizero}{$\pi_{0.5}$}
\newcommand{\env}{E}
\newcommand{\EnvCheck}{\textsc{Check}}
\newcommand{\EnvFree}{\textsc{Free}}
\newcommand{\ReHome}{\textsc{Home}}

\definecolor{cworld}{HTML}{C9302C}

\newif\ifdraftmarks
\draftmarksfalse

\begin{document}

\title{VLAs Execute, Code Recovers:\\3D Skill Envelopes Bridge the Compositional Gap\\in Agentic Manipulation}

\author{\paperauthorblock}

\makeatletter
\let\paper@titleblock\@maketitle
\def\@maketitle{\paper@titleblock
  \vspace{0.2\baselineskip}%
  \noindent\begin{minipage}{\textwidth}\centering
    \includegraphics[width=0.92\textwidth]{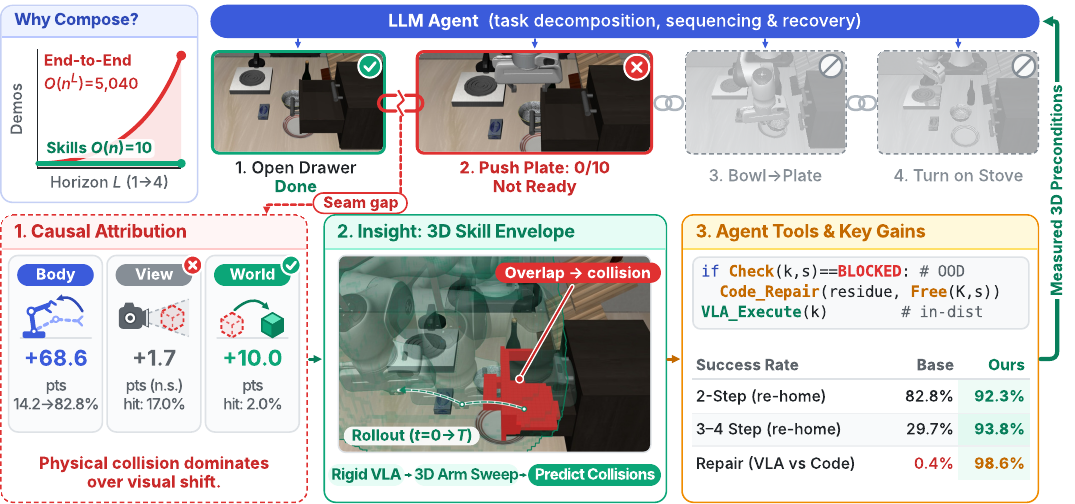}%
    \setcounter{figure}{0}\def\@captype{figure}
    \caption{\textbf{3D Skill Envelopes bridge the compositional gap between frozen VLAs and code recovery.} \emph{Top:} End-to-end demonstrations scale as $O(n^L)$, motivating $O(n)$ skill composition; yet after \emph{open drawer}, \emph{push plate} succeeds \ExHome{} even when re-homed, halting steps 3--4. (1)~Across \NStates{} handoffs, re-homing (\emph{Body}) fixes pose shift and hiding residue (\emph{View}) barely helps, while physically restoring residue (\emph{World}) recovers \DWorld{} pts and cuts collisions to \CWorld\%. (2)~Solo rollouts profile each skill's 3D \emph{Skill Envelope} (teal), flagging overlap (red) before execution. (3)~Via \EnvCheck{} and \EnvFree{}, concise code recovers blocked states (\PctPrim\% vs.\ \PctInv\% for inverse VLA prompting), enabling reliable multi-step composition.}%
    \label{fig:seam}%
  \end{minipage}}
\makeatother

\maketitle

\begin{abstract}
Composing frozen Vision-Language-Action (VLA) skills under an agentic planner promises scalable long-horizon manipulation without end-to-end demonstrations, yet strong atomic skills routinely fail at the seams between them even under a correct plan.
Standard remedies either reset the robot's pose or draft entry conditions in language, missing the physical cause we uncover through controlled causal interventions: even after re-homing, a frozen VLA rigidly retraces its habitual trajectory and collides with displaced scene residue, whether or not the residue is visible.
To address this, we propose the \emph{Skill Envelope}, a policy-measured 3D geometric precondition that captures the volume a frozen skill sweeps during its own successful executions.
Because the rigidity that makes a frozen skill fragile at seams also makes its trajectory predictable in advance, we expose each envelope as two classical 3D tools (\EnvCheck{} to flag blocking residue before execution and \EnvFree{} to find safe placements), establishing a division of labor where VLAs execute in-distribution skills and code recovers out-of-distribution states.
Across LIBERO-Goal (\pizero{}) and BOSS-44 (\texttt{BC-ViT-T}), Skill Envelopes outperform frontier VLMs at seam prediction (\ARQFiveOurs{} vs.\ \ARQFiveBestLang{} AUROC), lift autonomous LLM composition from \SPooledBThree\% to \SPooledBFour\% on LIBERO and from \SBossBThree\% to \SBossBFour\% on BOSS-44, and deliver a \XSpeedGated$\times$ speedup under live RGB-D gating.
Together, these results show that the compositional boundary between high-level agents and frozen visuomotor skills is physical, policy-specific, and measurable: grounding that boundary in 3D geometry rather than language simultaneously unlocks reliable multi-step autonomy and slashes LLM and VLM overhead.
Stage~2 will deploy this online RGB-D loop on physical hardware, scale to high-clutter benchmarks and a third policy family, and replace fixed primitives with envelope-verified code synthesis.
\end{abstract}

\IEEEpeerreviewmaketitle

\begin{figure*}[t]
\centering
\includegraphics[width=0.91\textwidth]{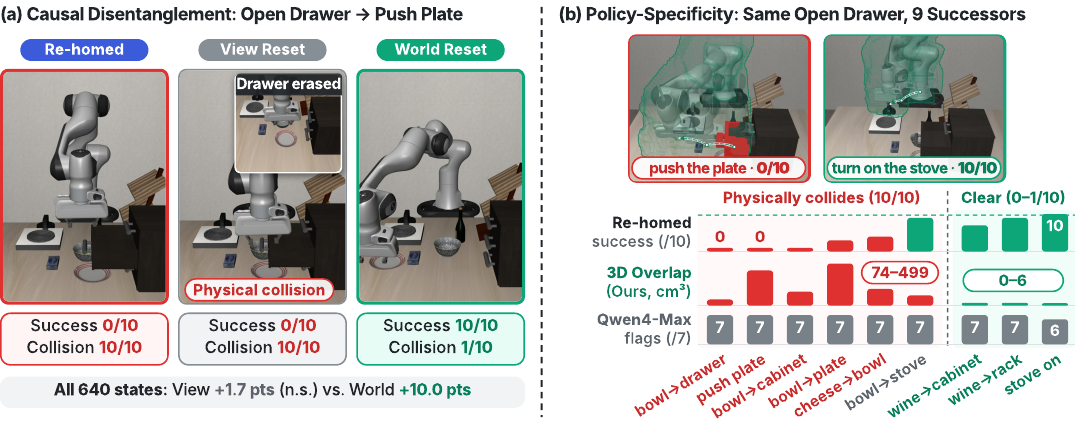}
\caption{\textbf{Seam failures are physical and policy-specific.} (a)~Matched-seed interventions on \emph{open middle drawer} $\to$ \emph{push plate} (10 trials): \emph{Re-homed} fails (\ExHome) and collides \ExHomeContact{}; erasing the drawer from pixels (\emph{View Reset}, inset) leaves success (\ExView) and collisions (\ExViewContact) unchanged; closing it (\emph{World Reset}) succeeds (\ExWorld). (b)~\emph{Top:} \emph{push plate}'s envelope hits the drawer (red, \ExHome) while \emph{turn on stove} sweeps clear (10/10). \emph{Bottom:} across all nine successors (dashed: World Reset 9--10/10), 3D overlap separates colliding from clear skills, whereas \texttt{Qwen4-Max} flags nearly every skill (7 held-out trials/pair).}
\label{fig:diagnosis}
\end{figure*}

\section{Introduction}
\label{sec:intro}

Everyday long-horizon manipulation cannot be solved by enumerating end-to-end demonstrations: even ten atomic skills yield 5{,}040 four-step permutations with distinct intermediate world states.
A natural alternative is \emph{agentic composition}: pairing a high-level LLM agent with a library of frozen, contact-rich Vision-Language-Action (VLA) skills~\cite{saycan,cap,baton,harnessvla}.

Yet strong atomic skills routinely collapse at the transitions (\emph{seams}) between them.
In LIBERO-Goal~\cite{libero}, a frozen \pizero{} policy~\cite{pi05} executes \emph{open the middle drawer} and \emph{push the plate to the front of the stove} reliably in isolation.
Chained directly, however, the second skill drops to \ExDirect{}; even after returning the arm to its starting pose (\emph{re-homing}), it remains at \ExHome{} because the arm slams straight into the open drawer (Fig.~\ref{fig:seam}).
Across \NPairs{} ordered skill pairs (\NStates{} handoffs), mean success falls from \SAtomic\% in isolation to \SDirect\% under direct chaining, recovering only to \SHome\% after re-homing: the predecessor is \emph{done}, yet the successor is \emph{not ready}.

Prior work recognizes this compositional gap, but guards seams through proxies rather than querying the policy itself.
Skill-chaining methods address only robot-pose mismatch while leaving displaced objects untouched~\cite{robocombine,causeway,atombridge}.
Agentic orchestrators build language transition rules through exploration~\cite{baton} or query VLMs to advance or retry~\cite{rui2026handoff}, paying a double penalty in fragile long-horizon reliability and minutes spent waiting on repeated LLM and VLM calls.
Post-hoc monitors instead wait for execution stalls and rely on learned recovery or policy fine-tuning~\cite{wagh2026seams}.
None of these asks the physical question directly: \emph{what physical precondition does a frozen policy actually require to succeed, and how can an agent check it before invocation?}

We answer this question in two steps.
First, controlled interventions on the exact same \NStates{} re-homed handoffs separate visual shift from physical world shift: erasing displaced objects (\emph{residue}) from pixels barely helps, whereas physically restoring them recovers most of the gap and slashes collisions (Fig.~\ref{fig:diagnosis}a).
Seam failure thus reflects \emph{physical, policy-specific hidden preconditions}: a frozen VLA rigidly replays its habitual sweep straight into obstructing residue.
Second, \emph{the rigidity that makes a frozen skill fragile at seams also makes its trajectory measurable in advance.}
Solo nominal rollouts, needing no demonstrations or failure data, record the 3D volume swept by the arm, gripper, and carried object as a \emph{Skill Envelope}.
Exposed via two lightweight 3D tools (\EnvCheck{} and \EnvFree{}), this envelope operationalizes a clean division of labor: \textbf{VLAs execute in-distribution, code recovers out-of-distribution states, and policy measurement, rather than language, draws the boundary.}

This Stage-1 paper makes three contributions:
{\makeatletter\@beginparpenalty=10000\makeatother
\begin{enumerate}
{\interlinepenalty=10000\item \textbf{Causal diagnosis of the compositional gap (Sec.~\ref{sec:diagnosis}):} We disentangle body, view, and world shift via controlled interventions on matched handoffs, revealing that residual failures after re-homing stem from physical, policy-specific collisions rather than visual shift.\par}
\item \textbf{3D Skill Envelopes for agentic composition (Sec.~\ref{sec:method}):} We introduce $O(n)$ envelope profiling from successful rollouts alone as a structured 3D geometric prior for frozen skills, exposed via \EnvCheck{} and \EnvFree{} as agentic tools pairing VLA execution with code recovery.
\item \textbf{Empirical validation of distribution-aware composition (Sec.~\ref{sec:results}):} We validate this division of labor across benchmarks, policy families, and LLM backends, showing that geometric readiness outperforms frontier VLMs, concise code far surpasses policy prompting for recovery, and live RGB-D gating jointly boosts multi-step reliability and slashes agentic orchestration latency.
\end{enumerate}}

\begin{figure*}[t]
\centering
\includegraphics[width=0.88\textwidth]{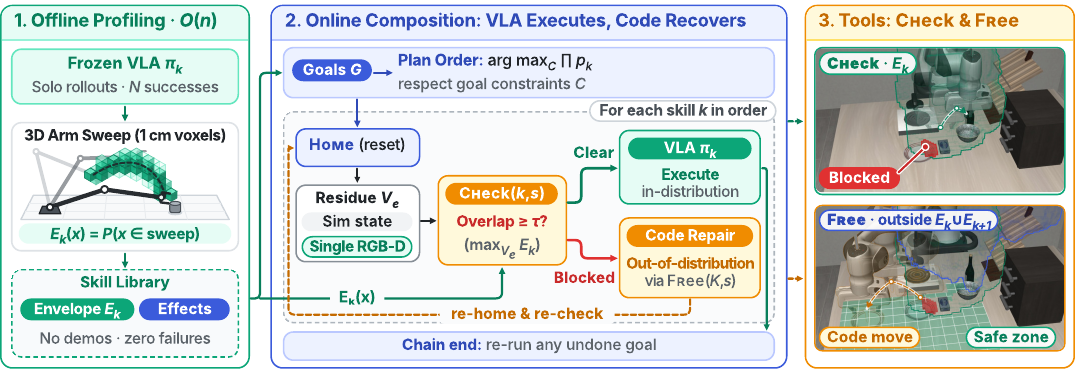}
\caption{\textbf{How 3D Skill Envelopes divide labor between frozen VLAs and code.} \emph{(1)~Offline Profiling ($O(n)$):} Solo rollouts of frozen $\pi_k$ voxelize swept trajectories into occupancy $\env_k(x)$ and log each skill's physical effects. \emph{(2)~Online Composition:} Goals $G$ are ordered under constraints $C$; at each seam, \ReHome{} resets the arm, state or RGB-D sensing builds residue $V_e$, and \EnvCheck{} tests $\env_k$ overlap, dispatching $\pi_k$ when clear or code repair (with \EnvFree{}) when blocked; undone persistent goals re-run at chain end. \emph{(3)~Tools:} On \emph{push plate} $\to$ \emph{cheese in bowl}, \EnvCheck{} flags the plate (red) in $\env_k$ (teal) and \EnvFree{} locates a safe zone outside $\env_k \cup \env_{k+1}$ for closed-loop relocation (amber arrow, RQ2).}
\label{fig:method}
\end{figure*}

\section{Why Skills Fail at Seams: A Causal Diagnosis}
\label{sec:diagnosis}

We evaluate frozen \pizero{} across LIBERO-Goal's ten skills, forming \NPairs{} ordered two-skill pairs with disjoint target objects (ten handoffs each, \NStates{} total).
Between predecessor $A$ and successor $B$, three factors shift out of $B$'s training distribution: the robot's \emph{body} pose, the camera \emph{view} of the scene, and the physical \emph{world} state altered by $A$.

\noindent\textbf{Physical collision dominates visual shift.}
We disentangle all three via controlled interventions on identical snapshotted handoffs replayed with matched action-sampling seeds (Fig.~\ref{fig:diagnosis}a).
Direct chaining collapses success from \SAtomic\% to \SDirect\%; \emph{re-homing} removes body shift and reaches \SHome\%, leaving a \DGapHome-point gap.
Applying \emph{View Reset} (hiding residue from the renderer while keeping physical colliders intact) yields a statistically insignificant \DView{}-point change and leaves collisions unchanged (\CView\% vs.\ \CHome\%; \ExViewContact{} on \emph{open middle drawer} $\to$ \emph{push plate}).
In contrast, \emph{World Reset} (restoring displaced objects to their nominal poses) recovers \DWorld{} points (to \SWorld\%) and cuts collisions to \CWorld\%.
While visual shift can hurt narrower policies~\cite{boss}, for a strong pretrained VLA in a shared workspace, failures after re-homing are governed by physical collision.

\noindent\textbf{Policy-specific, not scene-generic.}
In Fig.~\ref{fig:diagnosis}b, the same open middle drawer breaks \NBroken{} of nine successors (${\le}\,$\ExBrokenMax), leaves one mildly affected, and spares \NSpared{} (${\ge}\,$\ExSparedMin).
Whether residue breaks a seam cannot be judged from scene state or task language alone: it depends on the 3D corridor a frozen policy actually sweeps.

\noindent\textbf{Done is not ready.}
Finishing a predecessor satisfies its goal yet violates the successor's \emph{hidden preconditions}: never having seen such residue along its approach corridor, the frozen policy rigidly replays its habitual sweep into contact.
Because that corridor is repeatable, we can measure its swept volume in advance and flag obstructions before execution.

\section{Skill Envelopes and Distribution-Aware Composition}
\label{sec:method}

\subsection{Measuring the Skill Envelope}
\label{sec:envelope}

To measure a frozen skill $k$'s spatial requirements directly from its policy $\pi_k$ (Fig.~\ref{fig:method}, left), we roll out $\pi_k$ alone from nominal starts across $N$ successful episodes (\NEnvMin--\NEnvMax{} per skill).
At each step, forward kinematics (FK) maps the robot's links, gripper, and carried object onto a $1\,\mathrm{cm}$ voxel grid via their oriented bounding boxes (OBBs).
Letting $\mathrm{sweep}_i$ denote the set of voxels occupied during episode $i$, the empirical \emph{Skill Envelope} at voxel $x$ is:
\begin{equation}
\env_k(x) \;=\; \frac{1}{N}\sum_{i=1}^{N}\mathbf{1}\!\left[x\in \mathrm{sweep}_i\right]\;\in[0,1].
\label{eq:envelope}
\end{equation}
In a fixed workspace, $\env_k$ lives in the robot-base frame (and can be anchored to task-object frames under larger layout shifts).
Profiling requires \emph{no human demonstrations, failure rollouts, or symbolic models} and scales as $O(n)$ over $n$ skills rather than $O(n^2)$ over skill pairs: an envelope characterizes the successor regardless of which predecessor left the residue.
The same rollouts log each skill's \emph{effect} (displaced fixtures, objects, and end poses) for pre-execution ordering.

\subsection{Geometric Tools, Depth Sensing, and Readiness}
\label{sec:tools}

At a re-homed state $s$, residue $R(s)$ is the set of elements altered by prior skills (excluding the next skill's own targets).
Element $e$ occupies voxels $V_e$, overlapping $\env_k$ by peak probability $m_e$ and expected volume $o_e$ ($\mathrm{cm}^3$):
\begin{equation}
m_e(k,s)=\max_{x\in V_e}\env_k(x), \qquad o_e(k,s)=\sum_{x\in V_e}\env_k(x).
\label{eq:overlap}
\end{equation}
Intuitively, $m_e$ tests whether element $e$ enters a high-probability sweep voxel (a scale-free binary gate), while $o_e$ sums total overlap volume for continuous risk ranking.

\noindent\textbf{Dual sensing and 3D tools.}
With simulator access, $V_e$ comes from state differencing; without it, a single RGB-D frame after re-homing builds a unified residue $V_{\text{depth}}$ by back-projecting pixels $>1\,\mathrm{cm}$ closer than the start reference depth, masking out FK robot link OBBs (padded $2\,\mathrm{cm}$), and voxelizing onto the $1\,\mathrm{cm}$ grid without any detector or mesh.
Two classical 3D tools expose envelopes to the agent (Fig.~\ref{fig:method}, right):
\EnvCheck$(k, s)$ tests $m_e(k,s)\!\ge\!\tau$ ($\tau{=}0.05$ fixed a priori) over each residue element $e$ (or the unified $V_{\text{depth}}$) and returns blocking elements ranked by overlap;
\EnvFree$(K, s)$ computes, for a blocking free object $e$ given pending skills $K$, the nearest supported tabletop pose $\mathbf{p}$ whose shifted footprint $V_e(\mathbf{p})$ avoids both $\bigcup_{j\in K}\{x:\env_j(x)\!\ge\!\tau\}$ and scene occupancy.

\noindent\textbf{Training-free gating vs.\ calibrated readiness.}
Both collision scoring ($\sum_e o_e$) and \EnvCheck{} gating ($m_e\!\ge\!\tau$) require \emph{no failure data}.
When a calibrated probability $p_k(s)$ is needed for ordering or evaluation, we fit a logistic or 5-parameter hazard head ($b_0,b_1,b_2$; $a,c>0$) using sigmoid $\sigma(\cdot)$, overlap $(m_e, o_e)$, and re-homing error $(\tilde{e}_{\text{pos}}, \tilde{e}_{\text{ori}})$:
\begin{equation}
p_k(s)=\sigma(b_0{+}b_1\tilde{e}_{\text{pos}}{+}b_2\tilde{e}_{\text{ori}})\!\prod_{e\in R(s)}\!e^{-(a m_e + c o_e)},
\label{eq:readiness}
\end{equation}
multiplying survival across residues without object IDs.

\subsection{Distribution-Aware Division of Labor}
\label{sec:labor}

In our closed-loop pipeline (Fig.~\ref{fig:method}, center), semantic constraints $C$ over goals $G$ distinguish \emph{persistent end-state goals} (e.g., an open drawer required at termination, ordered last or re-invoked if undone) from \emph{transient sub-actions} (e.g., opening a drawer to place a bowl inside, where later closing keeps the bowl inside).
Before executing an $L$-step chain, planning selects the $C$-valid order $(k_1,\dots,k_L)$ maximizing $p_{\text{atomic}}(k_1)\prod\nolimits_{t=2}^{L}\hat{p}_{\text{rep}}(k_{t-1}\!\to\!k_t)$, where $p_{\text{atomic}}$ is a skill's isolated success rate and $\hat{p}_{\text{rep}}$ is the post-repair pairwise readiness predicted from profiled effects.

Before each skill $k$, \ReHome{} opens the gripper, lifts vertically to detach cleanly, and returns in joint space; sensing builds $V_e$ and calls \EnvCheck$(k,s)$.
On clear seams ($m_e\!<\!\tau$), the controller dispatches frozen $\pi_k$ in distribution.
On blocked seams, it invokes concise code repair, then re-homes and re-verifies: analytic impedance primitives close blocking drawers or hinged doors, while \EnvFree$(K,s)$ supplies collision-free targets to relocate blocking free objects.
At chain end, any undone persistent goal is re-invoked and verified against $G$.

\begin{table}[t]
\centering
\caption{Foundational validation across RQ1 and RQ2.}
\label{tab:quant}
\scriptsize
\setlength{\tabcolsep}{3.4pt}
\begin{tabular}{@{}lcccc@{}}
\toprule
\multicolumn{5}{@{}l}{\emph{(a) Generalization AUROC (RQ1): Envelope vs.\ baselines}} \\
Split (model, test set) & \multicolumn{2}{c}{Envelope} & Lookup & No-Geom \\
\midrule
Unseen pred.\ (logistic, \NStates) & \multicolumn{2}{c}{\textbf{\ALopo}} & \ALopoLookup & \ALopoNoGeom \\
Unseen pred.\ (5-param, \NStates{} fresh) & \multicolumn{2}{c}{\textbf{\AConfLopo}} & \AConfLopoLookup & -- \\
Cross-category, 0 drawers (5-param) & \multicolumn{2}{c}{\textbf{\AConfCross}} & \AConfCrossLookup & -- \\
\midrule
\multicolumn{5}{@{}l}{\emph{(b) Seam readiness vs.\ frontier VLMs (RQ2, \NRQFiveCommon{} held-out handoffs, AUROC)}} \\
Method (Img / Text) & \multicolumn{2}{c}{Overall} & Within-Pred. & Collision \\
\midrule
Envelope readiness (Ours) & \multicolumn{2}{c}{\textbf{\ARQFiveOurs}} & \textbf{\AStratOurs} & \textbf{\AContactEnvEval} \\
\texttt{Qwen4-Max} & \multicolumn{2}{c}{\ARQFiveBestLang{} / \ALangQwenMaxOne} & \AStratQwenMax & \AContactQwenMax \\
\texttt{Qwen3.8-27B} & \multicolumn{2}{c}{\ALangQwenLocTwo{} / \ALangQwenLocOne} & -- & -- \\
\texttt{GLM-5.3-Flash} & \multicolumn{2}{c}{\ALangGlmTwo{} / \ALangGlmOne} & \AStratGlmOne & -- \\
\midrule
\multicolumn{5}{@{}l}{\emph{(c) Two-step seam recovery success \% (RQ2: drawers and free objects)}} \\
Condition & All \NStates & Mid (\NMid) & Top (\NTop) & Plate (\NFreeFlag) \\
\midrule
Re-homing only & \SHome & \SMidHome & \STopHome & \SFreeFlagHome \\
Reactive / Sham$^*$ & \SJPosthoc & \SMidPosthoc & \STopPosthoc & \SFreeFlagSham \\
Proactive code (Ours)$^\dagger$ & \textbf{\SJRepair} & \textbf{\SMidRepair} & \textbf{\STopRepair} & \textbf{\SFreeFlagOurs} \\
World Reset (privileged) & \SJWorld & \SMidWorld & \STopWorld & \SFreeFlagWorld \\
\bottomrule
\multicolumn{5}{@{}p{0.96\columnwidth}@{}}{\vspace{1pt}\tiny $^*$Reactive repair on drawers; sham wait on plate. $^\dagger$\EnvCheck-gated (\NStates); proactive slide (\NNew); \EnvFree{} push (\NFreeFlag{} plates; all \NFreeTable{} tabletop: \SFreeTabHome\%$\to$\SFreeTabOurs\%, $p{=}\PFreeTabMcn$). 95\% CIs: RQ1 \DLopoLookupCI; RQ2 VLM \DRQFiveBestCI, drawer \DJRepairHomeCI, plate \DFreeFlagCI.}
\end{tabular}
\end{table}

This interface supports both deterministic planning (\emph{Plan+Fix}) and \emph{autonomous LLM orchestration}, where an LLM queries \texttt{forecast} (predicting pairwise ordering conflicts from profiled effects) and \EnvCheck{} (state or RGB-D); under \emph{envelope gating}, clear seams ($m_e\!<\!\tau$) dispatch $\pi_k$ directly without waking the LLM.

\section{Experiments}
\label{sec:results}

\begin{figure*}[t]
\centering
\includegraphics[width=0.83\textwidth]{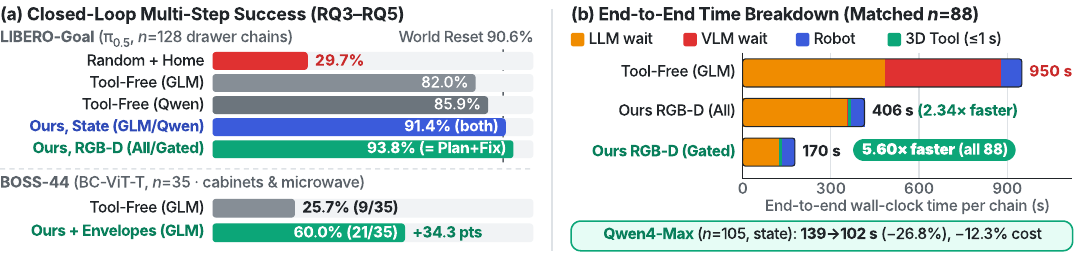}
\caption{\textbf{Closed-loop multi-step success and end-to-end latency (RQ3--RQ5).} (a)~Completion on LIBERO-Goal (\NChainDrawer{} chains, \pizero{}) and BOSS-44 (\NBossUnits{} held-out chains, \texttt{BC-ViT-T}): over tool-free VLM-judging agents (\emph{Tool-Free}), 3D envelopes lift \texttt{GLM-5.3-Flash} and \texttt{Qwen4-Max} to \SGlmBFour\%, reach \SRgbdBFour\% under live RGB-D (matching \emph{Plan+Fix}), and boost BOSS-44 by \DBossDiff{} pts ($p{=}\PBossMcn$). (b)~Time breakdown on matched succeeding chains ($n{=}\NRgbdMatched$, \texttt{GLM} RGB-D): querying the LLM at every seam without VLM polling (\emph{All}) runs \XSpeedSym$\times$ faster, while skipping it on clear seams (\emph{Gated}) achieves a \XSpeedGated$\times$ speedup across all \NRgbdMatched{} chains (bottom: matched \texttt{Qwen4-Max} state, $n{=}\NQwenMatched$).}
\label{fig:closedloop}
\end{figure*}

We organize our evaluation around five progressive research questions (RQ1--RQ5) spanning a pretrained VLA (\pizero{}) on LIBERO-Goal and a frozen Transformer-BC policy (\texttt{BC-ViT-T}) on BOSS-44 (Table~\ref{tab:quant}, Fig.~\ref{fig:closedloop}).

\subsection{Pre-Execution Detection and Generalization (RQ1)}
\label{sec:rq1}
We first ask whether training-free 3D overlap ($\sum_e o_e$) predicts collisions before execution and generalizes beyond memorized skill pairs.
Across \NStates{} LIBERO-Goal handoffs, state overlap achieves \AContact{} AUROC, and a single RGB-D difference after re-homing retains \ADepth{} AUROC while matching the state gate on all \DepthAgree{} open drawers.
On held-out splits (Table~\ref{tab:quant}a), our 5-parameter hazard head (Eq.~\ref{eq:readiness}) outperforms pair lookup across unseen predecessors, cross-category splits trained with zero drawers, and single- to multi-object residue transfer (\AConfSingle{} vs.\ \AConfSingleLookup), whereas removing geometric overlap collapses performance to chance.
Swept-volume overlap thus captures a general physical collision invariant rather than memorized pair statistics, and a single depth view suffices to measure it.

\subsection{VLMs vs.\ Geometry and Code Recovery (RQ2)}
\label{sec:rq2}
We next ask whether frontier VLMs can substitute for geometric readiness, and whether concise code should replace policy prompting for physical recovery.
Across three frontier VLMs and three prompt styles on \NRQFiveCommon{} held-out handoffs (Table~\ref{tab:quant}b), envelope readiness outperforms the strongest VLM by \DRQFiveBest{} AUROC and preserves strong within-predecessor discrimination (Fig.~\ref{fig:diagnosis}b).
Because a VLM observes only the scene rather than the policy's internal sweep, it flags nearly every successor behind an open drawer, whereas 3D geometry cleanly separates colliding from clear skills.

For recovery (Table~\ref{tab:quant}c), inverse language prompts to frozen \pizero{} close an open drawer in only \PctInv\% of attempts (\InvTotalFrac), whereas a ${\sim}30$-line slide primitive succeeds \PctPrim\% of the time, matching privileged World Reset overall while outperforming reactive repair on fresh drawers.
On blocking free objects, \EnvFree{} computes safe placements and lifts recovery across all \NFreeTable{} tabletop states to \SFreeTabOurs\%, approaching World Reset.
Elevated cabinet-top bottles (\NFreeWine{} states near reach limits) yield feasible plans in \FreeWinePlanPct\% of cases, marking a clear boundary: geometric clearing matches privileged resets whenever the obstacle lies within reach.

\subsection{Multi-Step Chains and Cross-LLM Orchestration (RQ3)}
\label{sec:rq3}
We next test whether seam gains compound on 3--4 step multi-goal chains where all goals must hold at termination (Fig.~\ref{fig:closedloop}a).
On \NChainDrawer{} drawer-bearing LIBERO chains, deterministic \emph{Plan+Fix} lifts completion from \SChainBOne\% (random order with re-homing) to \SChainBTwo\%, matching privileged World Reset and resolving two-drawer conflicts (\ChainTwoBOne{} to \ChainTwoBTwo).

Handing only goal text to autonomous agents across two LLM backends (\texttt{GLM-5.3-Flash} and \texttt{Qwen4-Max}), we compare a tool-free VLM-judging agent against an identical one with \texttt{forecast} and \EnvCheck{}.
Pooled across both backends ($n{=}\NPooledLLM$), envelope tools lift completion from \SPooledBThree\% to \SPooledBFour\% ($p{=}\PPooledMcn$), eliminate all end-of-chain re-invocations, and lower \texttt{Qwen4-Max} latency and cost (Fig.~\ref{fig:closedloop}).
Crucially, both LLMs converge to the exact same \SGlmBFour\% ceiling once given 3D tools: externalizing physical preconditions removes LLM reasoning as the compositional bottleneck, letting a lightweight model match a frontier backend without retries.

\subsection{Online RGB-D Loop and Wall-Clock Speedup (RQ4)}
\label{sec:rq4}
To drop simulator state entirely, we replace privileged checks with live single-view RGB-D \EnvCheck{} (\RgbdMsLow--\RgbdMsHigh\,ms) across all \NChainDrawer{} chains, comparing symmetric LLM invocation against \emph{envelope-gated} dispatch that skips the LLM on clear seams.
Both live RGB-D modes reach \SRgbdBFour\% completion, matching \emph{Plan+Fix} with \RgbdAgreeSeams{} state-gate agreement and beating the paired tool-free control ($p{<}10^{-5}$).
On matched succeeding chains (Fig.~\ref{fig:closedloop}b), skipping \RgbdSkipPct\% of seams as clear cuts VLM calls to zero and yields a \XSpeedGated$\times$ wall-clock speedup over the tool-free agent.
Because most seams are clear, geometric gating wakes the LLM only for physical conflicts, dissolving the trade-off between autonomy and speed.

\subsection{Cross-Benchmark, Cross-Policy, and Door Transfer (RQ5)}
\label{sec:rq5}
Finally, we test transfer beyond pretrained VLAs and drawers to a distinct visuomotor imitation-learning family: on BOSS-44~\cite{boss}, we evaluate a frozen Transformer-BC policy (\texttt{BC-ViT-T}) across four scenes spanning two cabinet styles and a hinged microwave door.
On \NBossUnits{} held-out starts with zero development overlap (Fig.~\ref{fig:closedloop}a), envelope tools lift autonomous completion from \SBossBThree\% to \SBossBFour\% ($p{=}\PBossMcn$), matching or improving every scene including the hinged microwave (\BossMicroBFour{} vs.\ \BossMicroBThree) while cutting repairs by \RepBossCut\% and eliminating end-of-chain re-invocations.
Skill Envelopes thus capture a general geometric property of frozen visuomotor execution rather than an artifact of one VLA architecture or fixture type.

\section{Stage-2 Plan and Open Questions}
\label{sec:plan}

Stage~1 validates Skill Envelopes in simulation across two policy families, two benchmarks, and 3--4 step chains with fixed recovery primitives, leaving three natural next steps that directly test our remaining assumptions.
First (\textbf{Q1}), moving beyond clean simulated depth, deploying our gated RGB-D loop on physical hardware will test whether depth differencing and padded OBBs survive specular dropouts, calibration drift, and contact compliance.
Second (\textbf{Q2}), scaling to 6--10 step cluttered BEHAVIOR-1K and BOSS Challenge~3 tasks alongside a second pretrained VLA (\texttt{OpenVLA-OFT}) will test safe-space saturation under joint envelopes $\bigcup_j \env_j$ and cross-VLA generality.
Third (\textbf{Q3}), replacing hand-written primitives with envelope-verified LLM code synthesis~\cite{aspire} will test whether 3D preconditions safely gate generated recovery code for unseen kinematics and where learned fallbacks are needed.

\section{Related Work}
\label{sec:related}

\noindent\textbf{Skill chaining and seam diagnosis.}
Prior work documents chaining drops~\cite{boss} and focuses mainly on robot-pose alignment~\cite{robocombine,causeway,atombridge}.
The closest work is Wagh et al.~\cite{wagh2026seams}, who diagnose scene seam failures post hoc via learned recovery or fine-tuning; we isolate body, view, and world shift to prevent collisions proactively without policy retraining.

\noindent\textbf{Agentic orchestration of frozen VLAs.}
Recent systems pair planners with frozen VLAs and code primitives~\cite{harnessvla}.
The closest to ours, BATON~\cite{baton}, learns language transition rules via test-time exploration and calls code to restore handoffs; we keep this split but replace language rules with \emph{policy-measured} 3D preconditions capturing each frozen skill's swept corridor.

\noindent\textbf{Geometric preconditions of visuomotor policies.}
StageCraft~\cite{stagecraft} mitigates obstructions via VLM queries over labeled rollouts.
Closest in geometric spirit, DR-LfD~\cite{drlfd} extracts swept volumes from human demonstrations for TAMP; Skill Envelopes instead profile the \emph{frozen policy itself}, requiring no demonstrations, failure data, or symbolic models.

\section{Conclusion}
\label{sec:discussion}

Through controlled causal interventions, we showed that after re-homing, frozen VLA composition fails not from visual distraction, but because policies rigidly replay their habitual 3D sweeps into physical residue.
Rather than relying on language proxies or post-hoc failure recovery, we turn this rigidity into an asset: profiling each skill's $O(n)$ \emph{Skill Envelope} from successful rollouts yields \EnvCheck{} and \EnvFree{} for live RGB-D verification, concise code repair, and geometric LLM gating.
More broadly, our work offers a clear design principle for agentic manipulation: let VLAs execute in-distribution, delegate out-of-distribution recovery to code, and let policy-measured 3D geometry, rather than language, draw the boundary, unlocking both reliable autonomy and fast execution.

\bibliographystyle{plainnat}
\bibliography{refs_arxiv,refs_manual}

\end{document}